# Developing a Street View Imagery Processing Pipeline for Building-Level Vacant House Assessment

Dongjoon Park[a], Sunjae Lee[b], Jung Woo Kim[c], Bumjoon Kang[a*]

[a]Department of Architecture and Architectural Engineering, Seoul National University, Gwanak-ro 1, Gwanak-gu, Seoul 08826, Republic of Korea

[b]Department of Urban Engineering, Graduate School of Engineering, The University of Tokyo, 7-3-1 Hongo, Bunkyo-ku, Tokyo, 113-8656, Japan

[c]Institute of Construction and Environmental Engineering, Seoul National University, Gwanak-ro 1, Gwanak-gu, Seoul 08826, Republic of Korea

[*]Corresponding Author

Email addresses: gbjun7333@snu.ac.kr (Dongjoon Park), leesunjae@g.ecc.u-tokyo.ac.jp (Sunjae Lee), w0016@snu.ac.kr (Jung Woo Kim), bumjoon@snu.ac.kr (Bumjoon Kang)

39-514 Gwanak-ro 1, Gwanak-gu

Seoul 08826, Republic of Korea

Department of Architecture & Architectural Engineering

Seoul National University

**Acknowledgements**

This work was supported by the National Research Foundation of Korea (NRF), funded by the Korean Government (MSIT) (RS-2025-00556250), via the Institute of Construction and Environmental Engineering at Seoul National University.

# Developing a Street View Imagery Processing Pipeline for Building-Level Vacant House Assessment

## Abstract

Street view imagery (SVI) is widely used to audit urban environments remotely. Automated assessments usually assume that each image shows the target building clearly. In practice, an image may lack task-relevant evidence or show a neighboring house. If this goes unchecked, a model's score cannot show whether it judged the correct building. We present a reproducible SVI processing pipeline that makes image suitability and target identity explicit and checkable. We apply it to vacant house assessment in South Korea, using 10,311 images of 1,609 detached-house parcels. Task-suitability filtering adapts its criteria to image composition and administrative area. This keeps coverage even across areas. While keeping as many images as a single uniform filter, our filtering left 101 more parcels with a usable image. Target-house alignment identifies the parcel's house from camera metadata and parcel boundaries. It needs no building footprints or additional training. Alignment matched human labels for 85.6% of images, compared with 79.8% for the best simple rule. In vacancy classification, filtering and alignment raised AUROC from 0.684 to 0.719, but marking a different house in the same image still gave 0.709. Grad-CAM showed that the model often looked at areas outside the target house. Image suitability, target identification, and model evidence are therefore separate problems, and scores alone do not resolve them. Recording which images were used and which house was selected lets reviewers verify each assessment and can support human virtual audits.

## 1. Introduction

Street view imagery (SVI) is now widely used to audit urban environments remotely, offering detailed, realistic views of the built environment across many urban areas (Biljecki & Ito, 2021; Steinmetz-Wood et al., 2019). A review found that 96 academic papers published from 2010 to 2023 used SVI to assess streets, sidewalks, traffic, or land-use patterns (Dai et al., 2024). Yet the abundance of available imagery does not guarantee that individual images contain evidence relevant to their analytical purpose. Uneven capture locations and revisit intervals (Kim & Jang, 2023), inconsistent image quality, limited metadata (Hou & Biljecki, 2022), and coverage bias (Fan et al., 2025) all affect analytical validity. Thus, SVI-based analysis depends not on image volume, but on what images capture and how they are collected.

A central challenge in SVI-based research is determining whether images suit the intended analysis. Because image suitability is task-dependent, its criteria vary with the phenomenon being evaluated and the analytical purpose (Devillers et al., 2007; Fan et al., 2025). Most existing assessments focus on technical availability criteria such as capture date, camera heading, field of view, metadata completeness, and occlusion (Hou & Biljecki, 2022; Kim & Jang, 2023). These criteria do not show whether an image contains the information needed for the intended assessment. For example, an image suitable for street-level greenery may not be adequate for assessing a particular house's physical condition (Li et al., 2015; Yan et al., 2024). Consistent model estimates also do not guarantee that the imagery adequately captures the target region of interest (Fan et al., 2025; Kim et al., 2021). To address this limitation, prior studies have proposed task-specific criteria. For building assessment, images should show the target building at an appropriate scale and angle and distinguish it clearly from neighboring buildings (Çelik & Sümer, 2020; Fan et al., 2025; Ogawa et al., 2021). Without appropriate SVI preprocessing, the target building may not be reliably identified, requiring separate visual identification (Roussel et al., 2024). Existing studies typically use SVI that captures both target buildings and their surroundings but rarely consider that building-level and contextual analyses may need different image characteristics, such as a clear view of a building's exterior or of the street around it (Kang et al., 2018; Li et al., 2023; Zou & Wang, 2021). Because image selection can affect model performance, poor performance cannot be attributed unambiguously to inadequate input images or to model limitations. Filtering criteria should therefore reflect what images need to capture for the intended assessment.

Recent studies have used SVI to assess housing vacancy (Yue et al., 2022; Zou & Wang, 2022). Vacant house assessment is a useful case to examine task-specific image suitability for three reasons. First, the task imposes strict image requirements. The target house must be correctly identified at the parcel level, and its exterior cues must be clearly visible. Second, vacant houses occur across contexts ranging from high-density metropolitan neighborhoods to low-density shrinking cities, where a house can appear very differently in SVI. Third, as vacant house identification has become a policy priority, official surveys in countries such as South Korea, Japan, and the United States rely on labor-intensive field inspections (Mallach, 2018; Ministry of Land, Infrastructure, Transport and Tourism, 2018; Sayuda et al., 2022), creating a strong need for scalable remote assessment. In South Korea's 2022–2024 surveys, field inspections found that nearly half of the properties initially identified as vacant from utility usage data were occupied (Ministry of Land, Infrastructure and Transport et al., 2025). This discrepancy shows that administrative records alone are insufficient to determine actual vacancy status and that additional verification is essential. SVI can be one means of this verification, allowing candidate properties to be screened before on-site inspection. Badland et al. (2010)

found that auditing neighborhood streetscapes with Google Street View took more than 20% less time than on-site audits.

This study proposes and evaluates a reproducible SVI processing pipeline for remote building assessment, using vacant house assessment as its empirical application. The pipeline comprises three related but separately evaluated procedures. First, we group images by visual features and administrative division, and evaluate task-suitability filtering by comparing group-specific filters with a uniform rule. We use the Building Capture Rate (BCR) to check whether the resulting image sets contain sufficient building evidence. Second, target-house alignment uses building instance masks, capture metadata, and parcel geometry to identify the surveyed house in each image. We assess alignment accuracy against human-verified target-house labels, compare it with simpler selection rules, and separately examine whether selections consistently identify the same house across viewpoints. Third, a classification experiment evaluates how image filtering and target-house alignment contribute to vacancy classification performance and the ability to review visual evidence associated with predictions.

## 2. Literature Review

### 2.1. Task suitability and image filtering

A field inspector observes the property itself, whereas an SVI-based analysis observes only a platform-supplied image, which may not provide enough information for the intended assessment. The task therefore determines what the image must reveal. Examples include screening for occlusion and unsuitable views in building assessment (Zeng et al., 2025), classifying images by the façade information each task requires (Wang, 2025), and filtering building views using visibility and image composition criteria (Liang et al., 2025).

In building assessment, a common threshold can have different effects depending on whether an image was taken in a dense urban block or a rural village, because building scale and occlusion vary with the surrounding environment. Liang et al. (2025) note that their filtering may exclude images from certain areas, such as those where dense greenery hides façades, and allow the filtering thresholds to be adjusted. For parcel-level tasks, where several views show one target, excluding all of them leaves no image for judging the parcel. Filtering therefore determines which images can be used and, with them, which properties remain observable.

### 2.2. Target-house identification and multi-view consistency

Sufficient visual information alone does not establish that the building shown is the surveyed house. Image-to-parcel matching is therefore needed to connect each image to the property it is meant to show. Errors in this matching can cause a property to be assessed with images of a different building. Existing studies have addressed this problem by coupling imagery with external spatial data. Ogawa et al. (2021) link street view images to building footprints, while OpenFACADES uses footprint-based visibility analysis to select views of each target building (Liang et al., 2025). In a related approach, O'Meara et al. (2025) use municipal curb-ramp coordinates and panorama metadata to delimit the image region where the target should appear before locating it at the pixel level. For building assessment, when several buildings fall within this region, establishing correspondence with the surveyed parcel requires selecting the target building from them.

Even when correctly linked to a parcel, each image captures only one viewpoint at one moment. Multiple viewpoints can reveal parts of a building obscured by vehicles, vegetation, or boundary walls. Lenjani et al. (2020) extract several views of a target building from nearby panoramas, and Ogawa et al. (2023) combine predictions from multiple images to estimate building attributes. When several buildings appear in images linked to the same parcel, the apparent target may shift between buildings across views. Evaluation therefore needs to establish whether the selected building is the surveyed house and whether selections across viewpoints consistently identify that same building.

### 2.3. Contextual cues and visual evidence

SVI-based vacancy assessment has used signs of neglect in both buildings and their surroundings. Zou and Wang (2021) combined scene-level classification with local patches showing deteriorated façades and unmaintained front-yard vegetation. In discussing misclassifications, they suggested that deteriorated patches from neighboring houses could cause a well-maintained target to be classified as abandoned. Their account illustrates that surrounding information may describe the surveyed property or introduce cues from another house. Because a model may base its prediction on features of neighboring buildings or the surrounding environment, correctly identifying the target does not explain the model's judgment of that house.

Visual explanation methods offer a way to examine image regions associated with predictions (Selvaraju et al., 2017). For example, Hu et al. (2023) combined Score-CAM with semantic segmentation in UPDExplainer to rank scene elements associated with urban physical disorder. They evaluated these rankings against human judgments, assessing the explanations alongside classification performance. For building assessment, reviewers can use the surveyed building as a reference to examine a prediction's visual explanation and distinguish highlighted regions on the target from those in its surroundings.

### 2.4. Research gap

Prior studies have released open-source frameworks that combine several SVI processing steps in a single workflow. ZenSVI supports workflows from image acquisition through analysis (Ito et al., 2025). OpenFACADES combines image selection, building extraction, and attribute prediction, and evaluates the resulting attributes and descriptions (Liang et al., 2025). Classification accuracy alone cannot show whether an image depicts the surveyed house. An image of a neighboring house may still yield a correct prediction if both share the same occupancy status. Few of the studies reviewed evaluate image suitability, target-house correspondence, and visual explanations together within the same parcel-level task. This study integrates all three components in a parcel-level assessment and evaluates image filtering, target-house correspondence, classification performance, and whether highlighted regions in correctly aligned images fall on the surveyed house. Evaluating them together can show whether a prediction succeeds or fails because of the image, the selected house, or the model. It can therefore confirm whether a difficulty in one procedure is separate from those in the others.

## 3. Data and Methodology

Figure 1 summarizes the pipeline. From the vacant house inventory, parcel polygons, and street view panoramas with their capture metadata (Figure 1a), it filters images for task suitability (Section 3.2), identifies the target house in each image (Section 3.3), and evaluates the resulting classification and its visual evidence (Section 3.4). To support reproducibility, the models, group-specific filtering thresholds, and alignment steps are described in the following subsections and the Supplementary Materials.

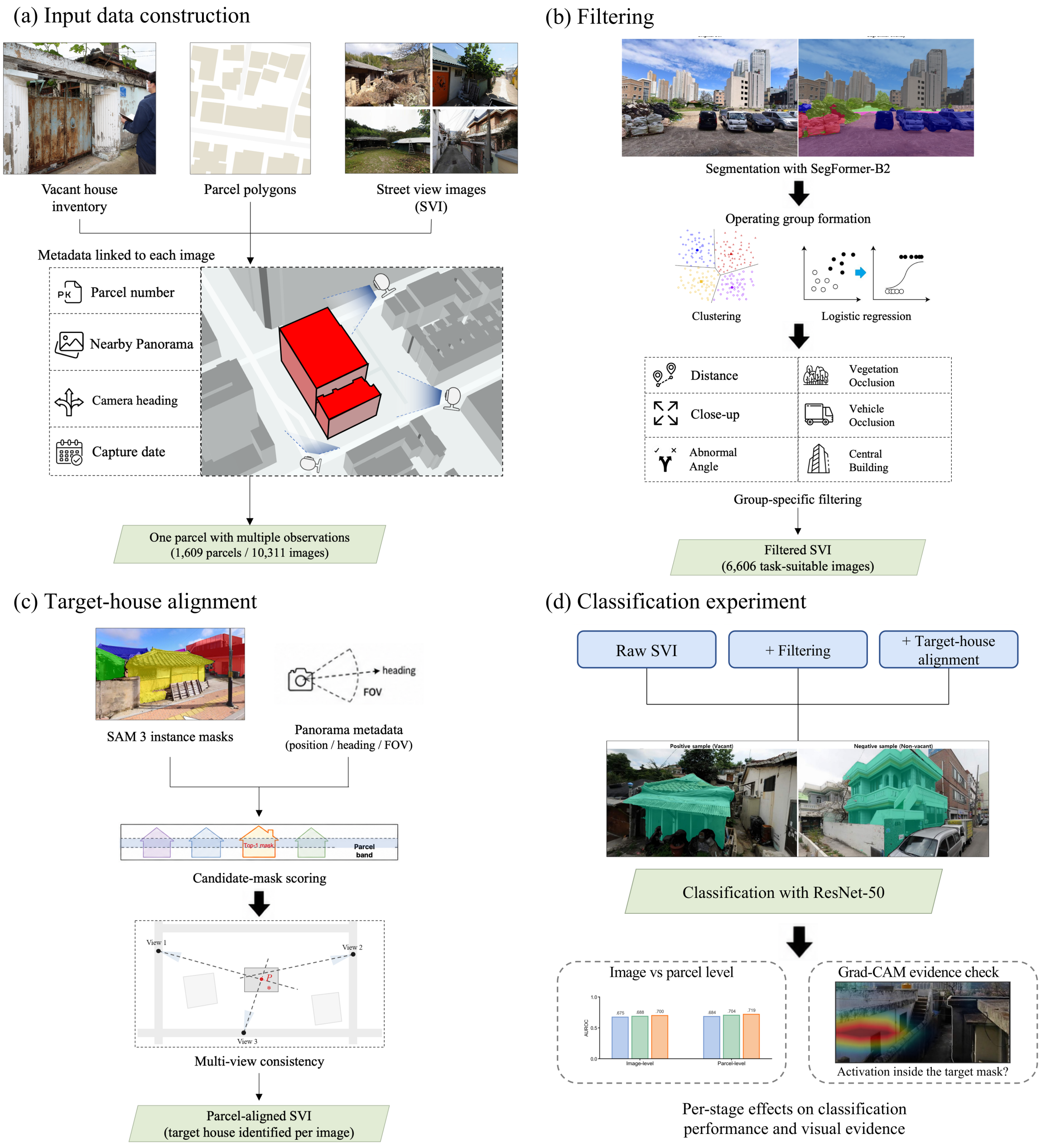

**Figure 1**. Overview of the proposed pipeline.

### 3.1. Data

This study draws on vacant house inventories released by local governments (*si*, *gun*, or *gu*) in South Korea. Since 2018, *the Act on Special Cases Concerning Unoccupied House or Small-Scale Housing Improvement* (2017) has required local governments to establish and maintain these inventories. We collected publicly available inventories between December 2022 and January 2026. Of the 226 municipalities nationwide, 103 published inventories with full addresses, and 33 of these provided parcel-level locations such as parcel numbers or coordinates, yielding 4,867 geocoded records. Each record included the vacancy determination, vacancy grade, property address, and, where available, survey date. We further restricted the analysis to detached houses because their individual exteriors can be reliably distinguished in SVI (Kang et al., 2018). Consequently, the final dataset is a purposive sample of administratively identified vacant houses that could be linked to parcel locations and SVI.

SVI was obtained from Naver Map and Kakao Map, South Korea's two primary SVI providers. From both platforms, we acquired the same capture metadata for each image (panorama position, camera heading, field of view, and capture date), on which the subsequent matching and target-house alignment rely. We matched each parcel number to its boundary in the Continuous Cadastral Map, identified nearby panorama locations, and retrieved images with the camera facing the parcel centroid. Each parcel was thus represented by images from different viewpoints, viewing directions, and capture dates.

For records with a survey date, we selected the images captured before and closest to that date to minimize the time between capture and survey. Otherwise, we selected the images closest to 2018, the year the Act took effect. We excluded records whose incomplete locations prevented linkage to cadastral and administrative data, as well as parcels without valid nearby panorama locations. The final dataset comprised 1,609 parcels and 10,311 images.

### 3.2. Task-suitability filtering

Filtering removes unsuitable images before modeling but is not designed to detect vacancy cues. Houses are hard to see for different reasons in different scenes. Vegetation may hide the exterior in one scene, whereas in another the house appears small and distant, so the two cases call for different filters. We therefore clustered images by scene and split each cluster into urban and rural images to form operating groups. We estimated each image's semantic composition with SegFormer-B2 fine-tuned on the 19 Cityscapes classes (Cordts et al., 2016; Xie et al., 2021). Following SVI studies that grouped street scenes by segmentation-derived view indices (Ito et al., 2025; Liang et al., 2023), we applied K-means clustering to the vegetation ratio, which indicates how green a scene is (Li et al., 2015), and the sky ratio, which indicates how open it is (Gong et al., 2018). We determined the number of clusters with two complementary criteria, the elbow method and the silhouette coefficient. We used both because the elbow method favors parsimony but often lacks a clear inflection point (Schubert, 2023), whereas the silhouette coefficient measures only cluster separation (Rousseeuw, 1987).

However, the vegetation and sky ratios do not capture the density and arrangement of surrounding buildings, which determine how large and how clearly a house can appear in the frame. In a dense urban block, houses fill much of the frame, whereas in a rural village the same camera setup captures far more road and sky (Fan et al., 2025). A single threshold tuned to one setting would therefore discard ordinary images from the other, but vacant houses occur in both. South Korea's sub-municipal administrative units distinguish urban neighborhoods (*dong*) from rural towns and townships (*eup* and *myeon*). The division is a common proxy for urban-rural context (Jeong et al., 2024) and is externally defined and independent of the imagery. We therefore classified each parcel as Urban (*dong*) or Rural (*eup* or *myeon*). To confirm that the division adds information beyond the image groups, we fitted logistic regression models predicting whether the building ratio met the building-pixel threshold and tested the division's contribution with a likelihood-ratio test (Hosmer et al., 2013). Based on these results, we combined the image groups and administrative divisions into operating groups.

We filtered images for six failure modes that make a house hard to see, covering its apparent size (distance, close-up capture), occlusion (vegetation, vehicle), and framing (abnormal angle, central building) (Hoiem et al., 2012; Hou & Biljecki, 2022). Filters were assigned to each operating group according to the visibility problems its scene type is prone to, and their thresholds were set to balance building evidence against image loss (Table 1). Supplementary Appendix C details each filter, with definitions and thresholds in Table C.1.

To quantify whether an image set contains sufficient building evidence, we define the Building Capture Rate (BCR). Let $r_i$ be the proportion of building pixels in image $i$. For threshold τ, BCR is defined as

$$BCR(\tau) = \frac{1}{N}\sum_{i=1}^{N} \mathbb{1}(r_i \geq \tau) \tag{1}$$

We computed BCR at $\tau = 1\%, 3\%, 5\%, \text{and } 10\%$, and direct inspection of the passing images supported τ = 5% as the principal criterion for whether a building had been detected. Supplementary Appendix B reports threshold-specific results.

We evaluated group-specific filtering, which adjusted filters and thresholds by operating group, against three baselines. Raw SVI retained all images, the building-ratio cut applied a single building-pixel criterion, and global filtering applied one common filter configuration. Except for Raw SVI, we calibrated the filtering conditions to similar overall retention so that comparisons reflected differences in strategy rather than in retention. BCR does not guarantee that the building is the surveyed house or that its vacancy can be judged from the image.

### 3.3. Target-house alignment

Even after filtering, images often show the target house alongside adjacent buildings, so a further step must establish which building corresponds to the surveyed parcel. Alignment proceeds in three steps, extracting building instance masks, scoring candidates with spatial metadata, and checking consistency across multiple views (Figure 2).

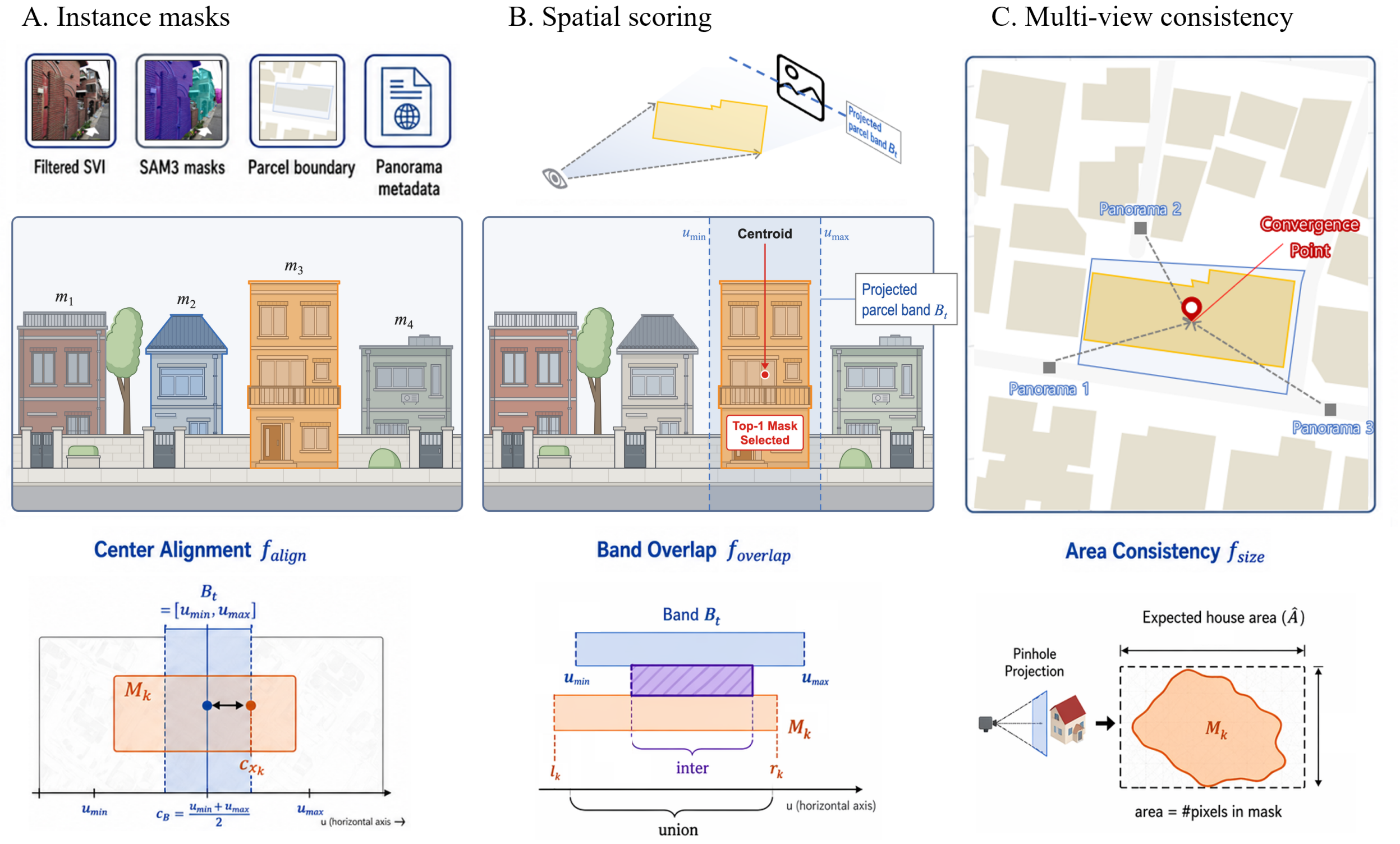

**Figure 2**. Target-house alignment workflow.

SAM 3 (Segment Anything Model 3) extracted building instance masks from each image (Carion et al., 2026). We then excluded images in which no house was detected, the number of instances was excessive, a boundary wall covered more than half of a mask, or the detected house was too small. Supplementary Table C.2 reports the threshold and calculation for each criterion.

We converted the parcel vertices into horizontal angular offsets from the camera heading and projected them onto equirectangular image coordinates. This projection defined the parcel band where the target parcel was expected to appear. Masks without horizontal overlap with the parcel band were removed. We scored each remaining mask on three spatial factors normalized between 0 and 1, namely center alignment within the band, horizontal overlap with the band, and area consistency relative to capture distance. We compared each factor alone and their product (the combined score) and adopted the best-performing criterion for all subsequent analyses.

The research team labeled the surveyed house in each image by matching parcel locations to surrounding panoramas. Ambiguous cases were flagged for joint review, and only confirmed labels were retained. We measured top-1 agreement as the proportion of images in which the selected mask matched the human label. Using the same evaluation set, we compared the spatial scores with four heuristic baselines (Random, Most-central, Nearest, and Largest).

Because images of the same parcel were captured under different conditions, their top-1 masks did not necessarily identify the same building. We therefore defined a sight line from each capture position through the center of its top-1 mask. For each parcel, we estimated the convergence point minimizing the squared perpendicular distances to these lines (Figure 3). We assessed consistency by the mean residual across viewpoints and the convergence point's distance from the parcel. Supplementary Appendix D provides the formulas and threshold rationale.

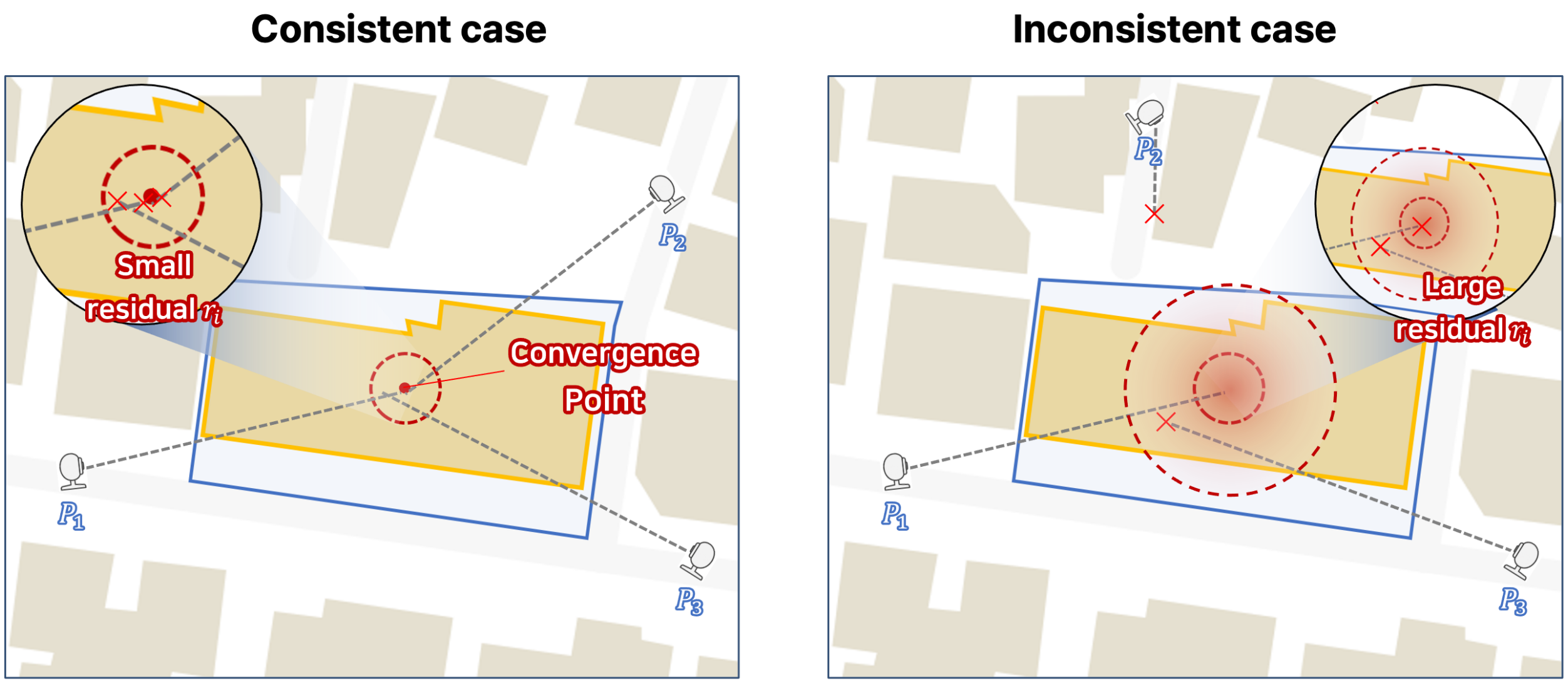

**Figure 3**. Multi-view consistency check via sight-line convergence.

### 3.4. Experimental setup

For binary classification, parcels registered as vacant served as positives and detached-house parcels not registered as vacant as negatives. To reduce confounding from neighborhood context, each positive parcel was paired with one negative inside the same administrative boundary, located within 50 m (López Ochoa & Zhai, 2025) but at least 10 m away (Baylis & Boomhower, 2026) so that the pair would not share a street scene. Eligible negatives had also undergone no administrative change during the preceding ten years and were processed with the same pipeline and capture-date rule as their paired positives.

To isolate the contributions of filtering and alignment, the classification experiment compared three sequential input conditions, namely Raw SVI, group-specific filtering, and target-house alignment after filtering. The final analysis used the 1,103 pairs available under all three conditions. Each parcel contributed its post-alignment image count to every condition, so that parcel-level results did not depend on image counts. To minimize spatial leakage, we kept each pair and all its images in a single split. We divided the pairs into five cross-validation folds (70% training, 10% validation, 20% test), with each pair tested once.

Because vacant house samples were limited, we fine-tuned a ResNet-50 classifier (He et al., 2016) initialized with ImageNet weights (Russakovsky et al., 2015). We trained models at the image level and evaluated them at the image and parcel levels with F1 and AUROC, reported as mean ± standard deviation across the five folds. Under the target-house alignment condition, the input image carried the top-1 mask as a fourth channel (Sun et al., 2024), while the Raw SVI and group-specific filtering conditions used the unmodified image. We marked rather than cropped the target so that contextual cues around the house remained visible and the input format stayed comparable across conditions. Two controls tested whether the channel helped by marking the target house. The shuffled-mask control used another house's mask from the same image, and the mask-only control used the target mask without the image. Finally, Grad-CAM, computed with a model not trained on the image, was used to visualize activated regions and determine whether predictions relied on the target house or the surrounding context (Selvaraju et al., 2017). Separately, as an exploratory test of cross-platform transferability, we ran the pipeline through target-house alignment on Google Street View images of Detroit (Supplementary Appendix A, Figure A.1).

# 4. Results

## 4.1. Sample and pipeline

Of the 10,311 images from 1,609 parcels, group-specific filtering retained 6,606, and target-house alignment was evaluated on 3,869 filtered images of vacant parcels. The classification experiment used the 1,103 vacant–non-vacant pairs with images under all three input conditions, so every condition covered the same 2,206 parcels.

## 4.2. Operating groups for task-suitability filtering

K-means clustering across all 10,311 images yielded three clusters. For values of $k$ from 2 to 8, the silhouette coefficient peaked at 0.545 when $k = 3$, the elbow curve leveled off beyond $k = 3$, and cluster assignments remained consistent across 20 random seeds. The three clusters were labeled Built-up, Green, and Open-sky. In Built-up images (N = 5,923, 57.4%), buildings occupied a large share of the frame. In Green images (N = 1,877, 18.2%), vegetation often obscured building exteriors. Open-sky images (N = 2,511, 24.4%) had unobstructed views, but buildings were captured from a distance. Figure 4 shows the proportion of images in each group. The image-based vegetation ratio was significantly correlated with satellite-based green cover (Supplementary Figure E.1).

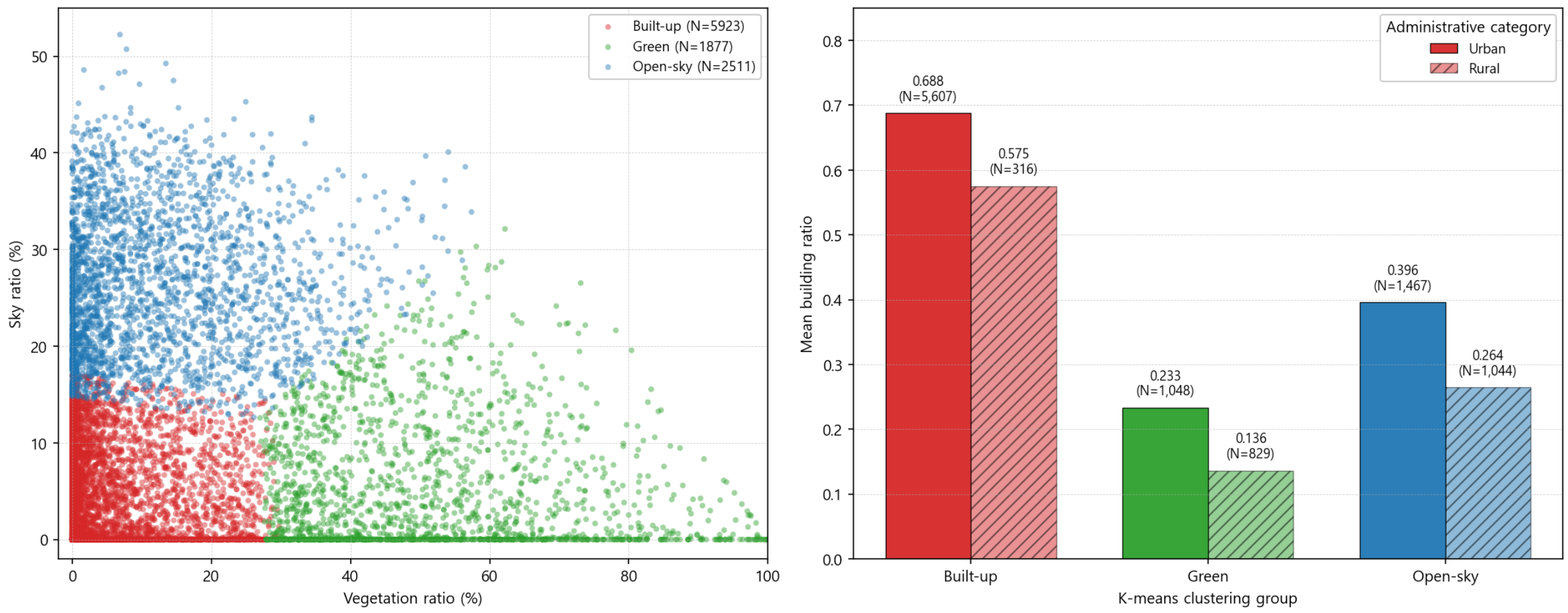

**Figure 4**. Construction of operating groups based on image composition (left) and administrative division (right).

Adding administrative division to the baseline logistic regression model, which included only the image groups, significantly improved model fit ($LR\ \chi^2(1) = 51.7, p < 0.001$). After controlling for image group, the odds that the building ratio met the threshold in Urban images were approximately 1.95 times those in Rural images (95% CI [1.62, 2.34]). We therefore adopted the six operating groups as the unit of analysis for filtering.

### 4.3. Task-suitability filtering performance

Group-specific filtering retained 6,606 of the 10,311 images (64.1%), with retention ranging from 86.8% in Open-sky × Urban to 26.2% in Green × Rural (Table 1).

**Table 1**. Filter configurations, thresholds, and retention for the six operating groups. Filters are coded as A = distance, B = vegetation occlusion, C = close-up, D = vehicle occlusion, E = abnormal angle, and F = central building.

| Operating group | Applied filters | Thresholds | Passed images | Retention |
|---|---|---|---|---|
| Built-up × Urban | A, C, F | $A \leq 30m$, $C \leq 0.85$, $F \geq 0.03$ | 3,615 | 64.5% |
| Built-up × Rural | A, C, F | $A \leq 30m$, $C \leq 0.90$, $F \geq 0.03$ | 247 | 78.2% |
| Open-sky × Urban | A, E*, F | $A \leq 30m$, E, $F \geq 0.15$ | 1,274 | 86.8% |
| Open-sky × Rural | A, E*, F | $A \leq 30m$, E, $F \geq 0.15$ | 848 | 81.2% |
| Green × Urban | A, B, D, F | $A \leq 30m$, $B \leq 0.50$, $D < 0.02$, $F \geq 0.05$ | 405 | 38.6% |
| Green × Rural | A, B, F | $A \leq 30m$, $B \leq 0.50$, $F \geq 0.10$ | 217 | 26.2% |
| Total | — | — | 6,606 | 64.1% |

**Filter E discards an image if building < 0.50, sky < 0.01, and road < 0.01.*

Before filtering, BCR ($\tau = 5\%$) ranged from 98.4% to 98.9% in the Built-up groups and from 93.6% to 97.1% in the Open-sky groups, whereas the Green groups ranged from 69.0% to 80.7%, up to approximately 30 percentage points lower. After filtering, BCR converged to 96.9–100.0% across all groups. The largest gains occurred in Green × Rural (+28.7 percentage points) and Green × Urban (+18.5 percentage points), which had the two lowest starting values (Figure 5). The vegetation-occlusion filter was the main driver of improvement in the Green groups, and the central-building filter in the Open-sky groups.

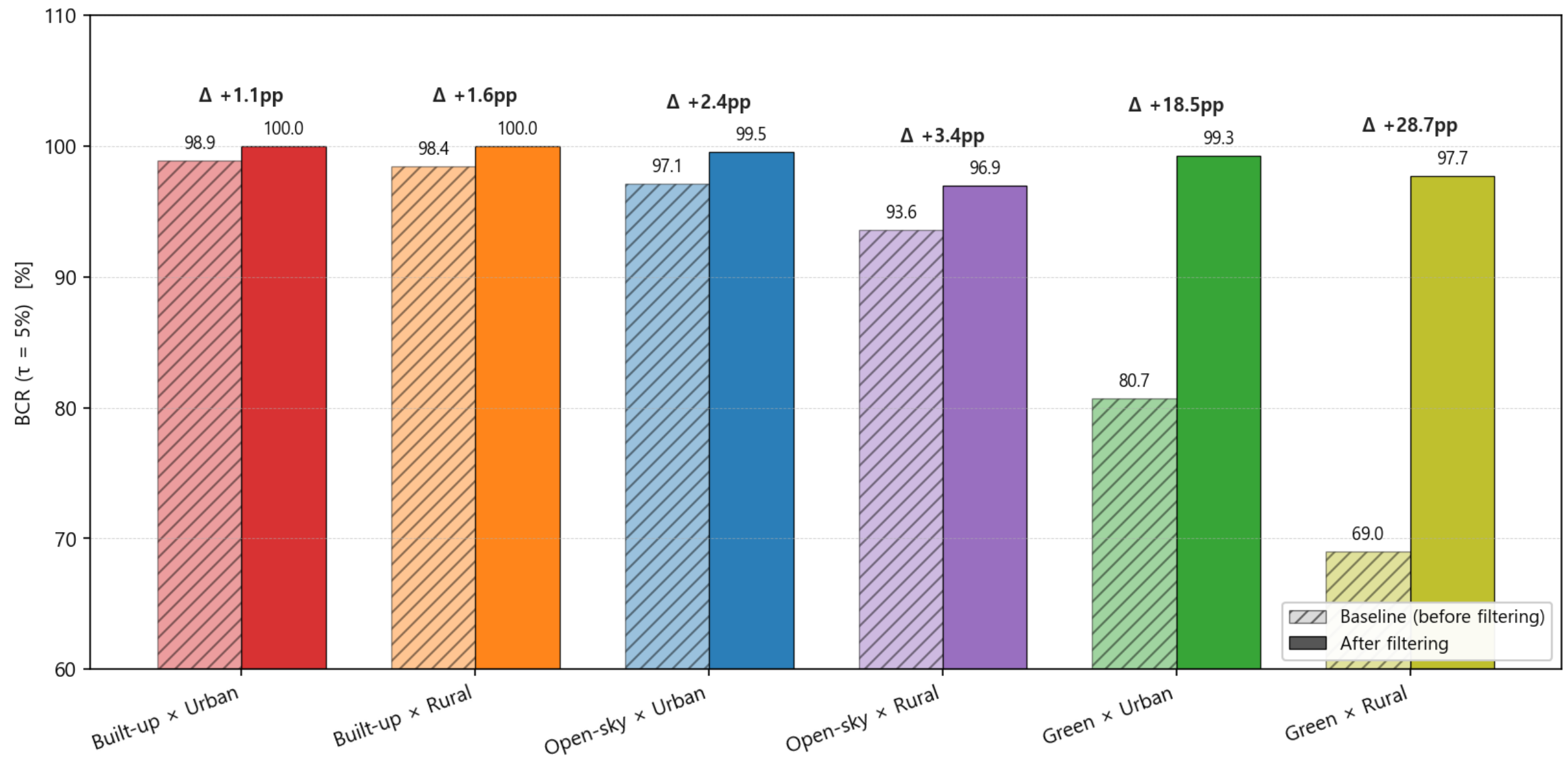

**Figure 5**. Building Capture Rate (BCR, τ = 5%) before and after group-specific filtering for the six operating groups.

Table 2 compares the four filtering conditions across groups. Under the building-ratio cut, the minimum group retention fell to 6.5%, and the disparity among groups reached 83.8 percentage points. Global filtering still left a minimum group retention of only 10.3%. Group-specific filtering more than doubled the minimum group retention and reduced the disparity to 60.7 percentage points. It also maintained an overall BCR of 99.4% and lost the fewest parcels (160, 9.9%).

**Table 2**. Retention, BCR, and parcel loss under four filtering conditions.

| Filtering condition | Overall retention | Min. group retention | Max. group retention | Retention disparity | Overall BCR | Parcels lost |
|---|---|---|---|---|---|---|
| Raw SVI | 100.0% | 100.0% | 100.0% | 0.0% | 93.8% | 0 (0.0%) |
| Building-ratio cut | 64.1% | 6.5% | 90.3% | 83.8% | 100.0%* | 261 (16.2%) |
| Global filtering | 62.6%** | 10.3% | 91.5% | 81.3% | 99.3% | 185 (11.5%) |
| Group-specific filtering | 64.1% | 26.2% | 86.8% | 60.7% | 99.4% | 160 (9.9%) |

**100% by construction, since the building ratio cut applies the BCR criterion itself.*
***Closest attainable retention to the 64.1% target.*

### 4.4. Target-house alignment performance

We applied the alignment procedure to the filtered images of vacant parcels. After excluding images with no candidate masks or with masks too small, 4,947 images remained. Images contained 1–8 candidate masks (about 67% had at least three). We further removed 652 images without a human-identifiable target, 423 without a parcel polygon in the cadastral map, and 3 whose panorama lay behind the parcel, leaving 3,869 for evaluation.

Figure 6 reports top-1 agreement for each selection criterion. Overlap with the parcel band performed best as a single criterion, reaching 85.6%. The four heuristic baselines (Random 47.8%, Most-central 68.1%, Nearest 77.5%, Largest 79.8%), center alignment (75.2%), and area consistency (63.1%) all fell below overlap. The combined score reached 83.9%, above all four baselines but below overlap alone.

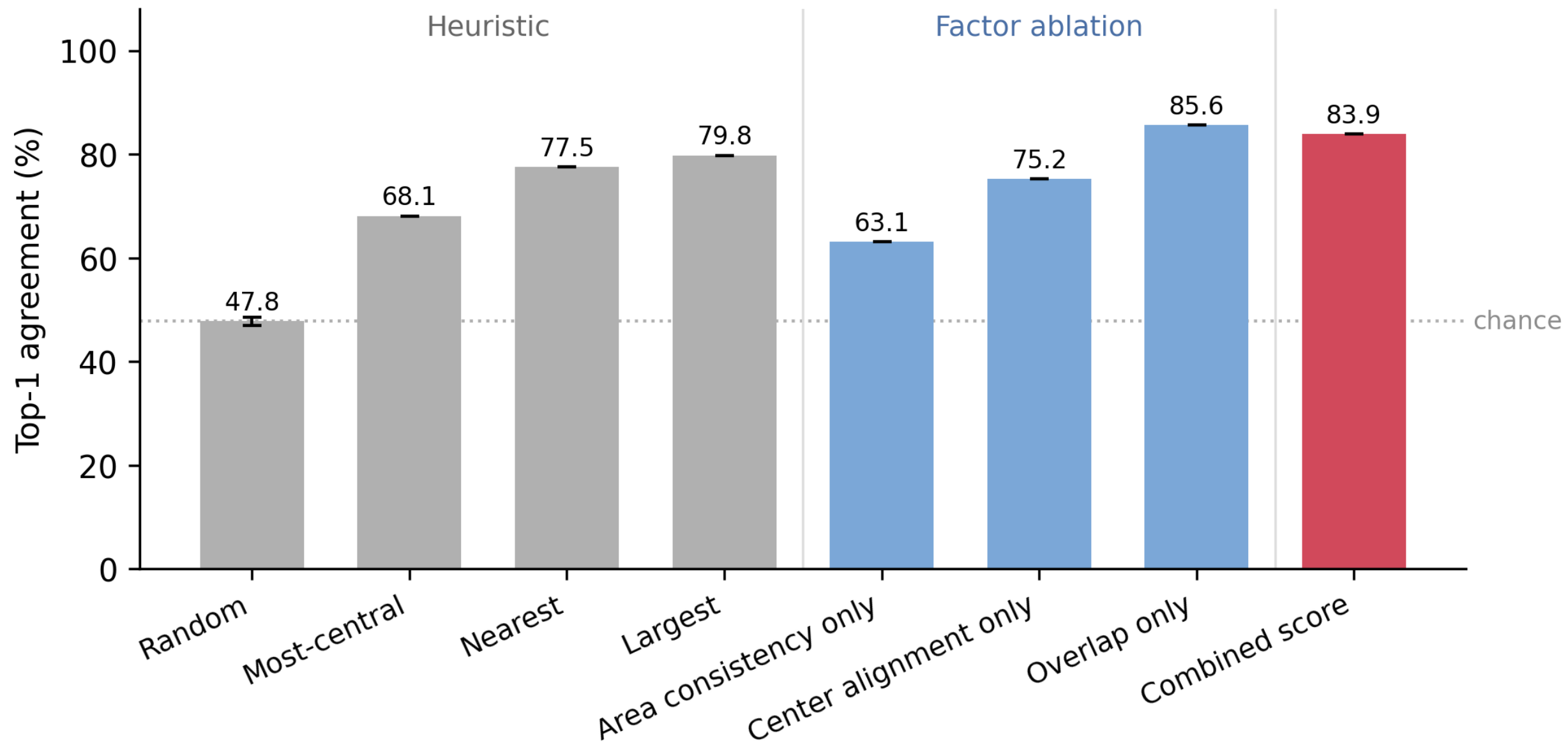

**Figure 6**. Top-1 agreement of heuristic baselines, single-factor scores, and the combined score.

The advantage of overlap grew with the number of candidates (Figure 7). For images with four or more candidates, area consistency and center alignment fell to 50–65%, whereas overlap still achieved 79.8%, above Largest (72.6%) and Nearest (70.5%), and its top-3 recall reached 98.1%. For three-mask images, the largest sample bin, top-1 agreement ranged from 78.3% (Built-up × Urban) to 89.7% (Green × Rural), so the Green groups combined the lowest filtering retention with the highest alignment agreement.

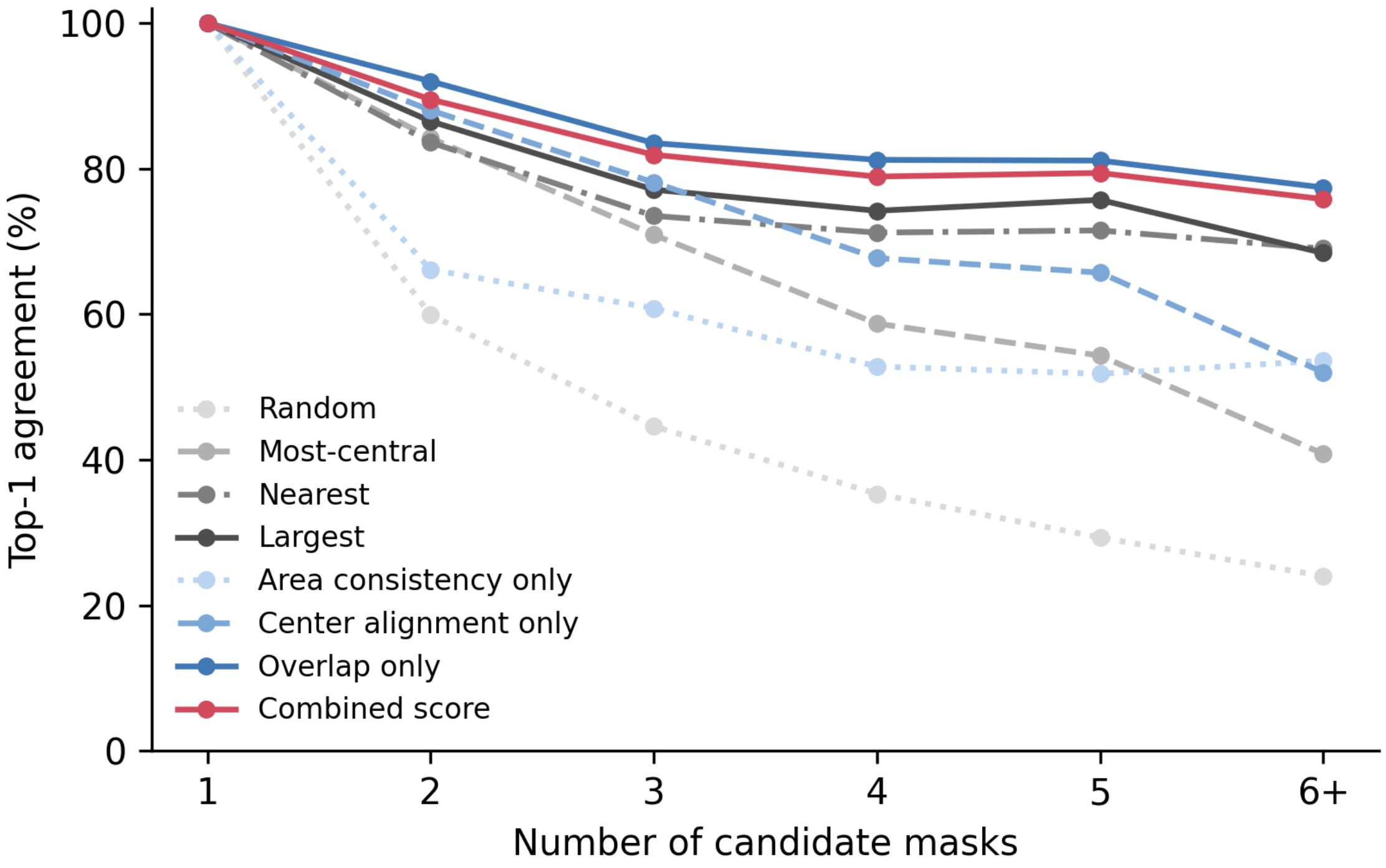

**Figure 7**. Top-1 agreement by number of candidate masks per image.

We assessed multi-view consistency for the 726 parcels with at least three viewpoints (Figure 8). Mean sight-line residuals were 3 m or less for 94.8% of parcels (688) and 5 m or less for 99.6% (723). Of the 630 parcels whose convergence point had a calculable parcel distance, 90.3% (569) satisfied both a residual of 3 m or less and a parcel distance of 10 m or less. Most masks selected from different viewpoints thus converged on one point near the target parcel.

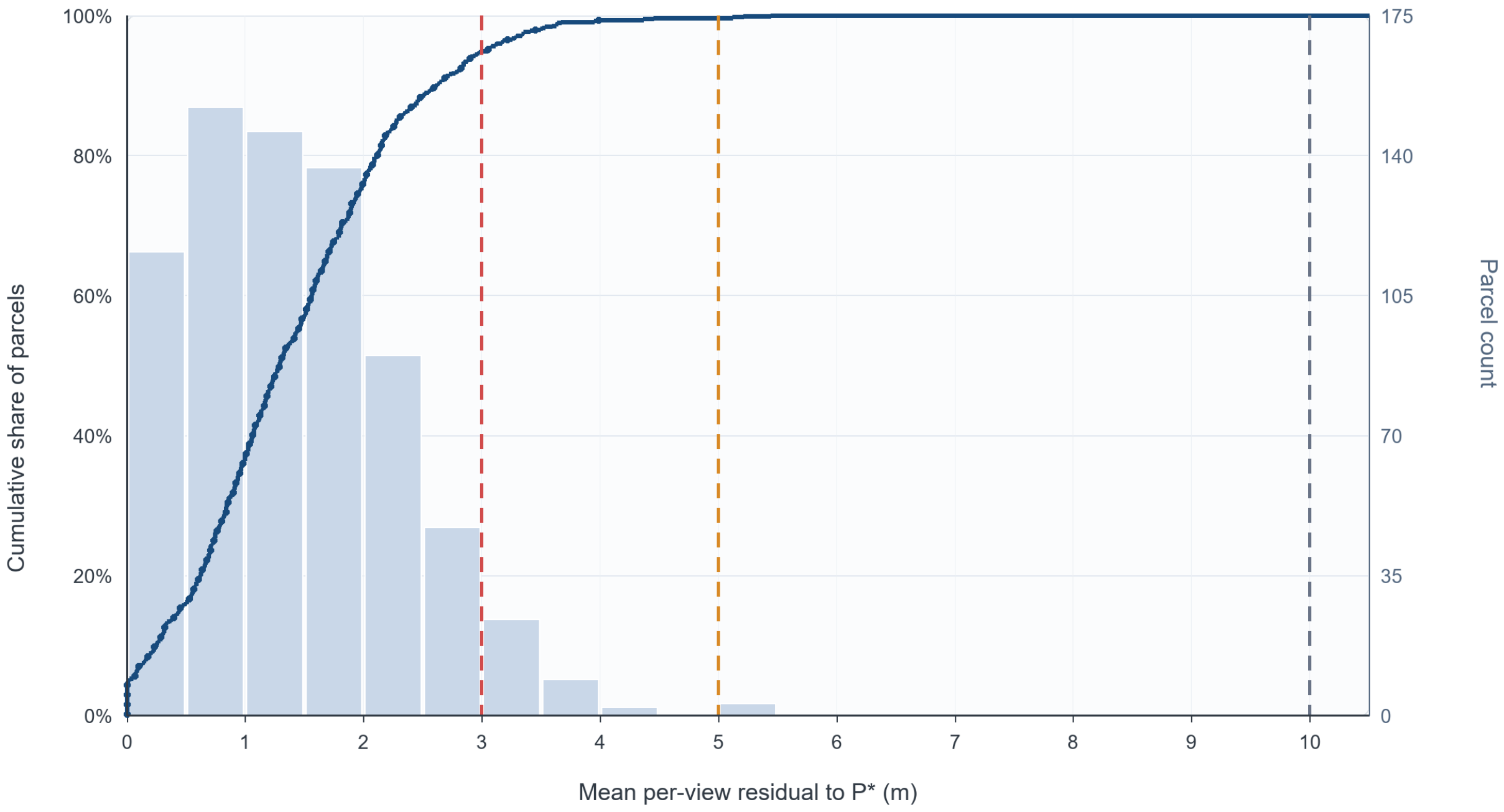

**Figure 8**. Distribution of multi-view convergence residuals.

*Dashed lines at 3 m, 5 m, and 10 m correspond to 94.8%, 99.6%, and 100.0% of parcels, respectively.*

### 4.5. Classification performance and visual evidence

Table 3 summarizes classification performance under the three input conditions and two controls. Because all conditions used identical parcels, splits, and model architecture, they differed only in the input images and the mask channel. Within each condition, parcel-level scores exceeded image-level scores by about 0.05 in F1 and by up to 0.02 in AUROC.

**Table 3**. Image-level and parcel-level vacancy classification performance for the three input conditions and two controls.

| Input condition | Image-level F1 | Image-level AUROC | Parcel-level F1 | Parcel-level AUROC |
|---|---|---|---|---|
| Raw SVI | 0.632 ± 0.026 | 0.675 ± 0.033 | 0.687 ± 0.023 | 0.684 ± 0.045 |
| + Group-specific filtering | 0.635 ± 0.024 | 0.688 ± 0.012 | 0.695 ± 0.010 | 0.704 ± 0.008 |
| + Target-house alignment | 0.650 ± 0.015 | 0.700 ± 0.011 | 0.698 ± 0.020 | 0.719 ± 0.018 |
| Shuffled-mask control | 0.643 ± 0.017 | 0.698 ± 0.008 | 0.688 ± 0.025 | 0.709 ± 0.006 |
| Mask-only control | 0.648 ± 0.014 | 0.671 ± 0.015 | 0.700 ± 0.010 | 0.686 ± 0.031 |

*(mean ± SD over five cross-validation folds, with F1 thresholds tuned on each fold's validation split.)*

Differences among the input conditions were largest for parcel-level AUROC, at 0.684 for Raw SVI, 0.704 with group-specific filtering, and 0.719 with target-house alignment. The shuffled-mask control reached 0.709, within 0.010 of alignment, and the mask-only control reached 0.686, close to Raw SVI. Parcel-level F1, in contrast, varied only from 0.687 to 0.700.

For correctly aligned images, the model trained with target-house alignment placed an average of 59% of its top 10% Grad-CAM-activated pixels within the human-labeled target mask, approximately 1.5 times the mask's mean share of the frame (41%). The model trained on filtered images placed 58% there, so the mask channel did not shift activation toward the target. Yet in more than one-third of the images, more than half of the most strongly activated pixels lay outside the target house, and in extreme cases all of them did. Activations outside the target appeared over adjacent buildings and other parts of the street scene (Figure 9).

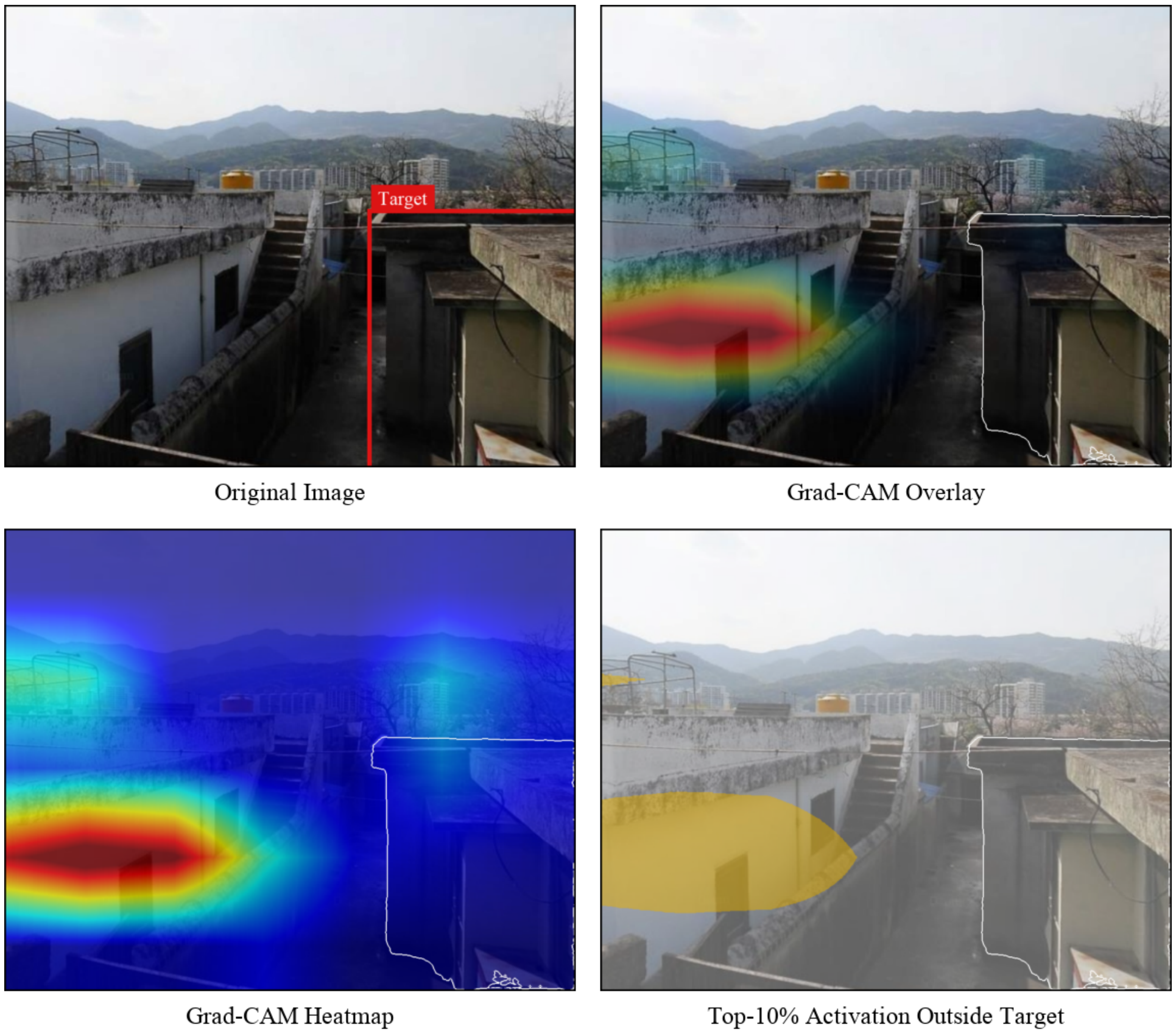

**Figure 9**. Grad-CAM activations versus the aligned target-house mask.

## 5. Discussion

Starting from 1,609 parcels (10,311 images), the pipeline applied task-suitability filtering and then target-house alignment. On 3,869 human-labeled images of vacant parcels, the selected house matched the label in 85.6% of cases. In the group with the lowest pre-filtering BCR, about three in ten images contained insufficient building-pixel information. In roughly one in three images evaluated for alignment, the most central building in the frame was not the surveyed house. A parcel processed without these checks can therefore be represented by an image with little building information or by a neighboring house. For individual building assessments, the image and its target need to be checked alongside the prediction.

### 5.1. Target-house alignment and existing pipelines

The most central building in a frame is an unreliable guide to the surveyed house, because each image was centered on the parcel centroid, which need not be where the house stands. Given building instance masks, the overlap rule reduces this mismatch by projecting the parcel boundary into each image, without additional model training. Making the selected building explicit lets a reviewer check it against the surveyed parcel. Sight lines through the selected masks converged for most parcels with three or more viewpoints, and the few parcels whose sight lines diverge can be flagged for review.

Earlier approaches located the target with building footprints (Liang et al., 2025; Ogawa et al., 2021) or with recorded target-object coordinates (O'Meara et al., 2025). Our rule replaces both with the parcel records and capture metadata already collected for image retrieval, so a target can be selected for any parcel on a cadastral map, even without a building footprint or recorded target coordinate.

Because target selection was evaluated separately from classification, against human labels and across viewpoints, the alignment and multi-view checks can be used alongside frameworks such as ZenSVI (Ito et al., 2025) and OpenFACADES (Liang et al., 2025). The criteria and intermediate outputs of each procedure are also documented, so the pipeline can be repeated in other regions.

## 5.2. Task suitability and parcel retention

At the same overall image retention, the building-ratio cut left 261 parcels without a retained image. As Table 2 shows, group-specific filtering reduced this number to 160 while maintaining an overall BCR of 99.4%. A common criterion applied to images taken in different places can thus remove every view of a property even when the remaining images reach a high BCR. When filtering results are reported, parcels with no retained image should be listed separately from those represented by at least one view, so that it is clear which surveyed properties remain in the observed sample.

Liang et al. (2025) allowed thresholds to be adjusted for different urban contexts. Our results show what that choice changes at the parcel level. Group-specific filtering left 101 fewer surveyed parcels without a retained image than the building-ratio cut, so this choice decides which surveyed properties can be assessed at all.

## 5.3. Model evidence and human verification

As Table 3 shows, classification scores differed only modestly across the input conditions, and marking another house in the same image scored almost the same as marking the target house. Label-agreement scores are therefore largely insensitive to whether the marked house is the right one. Evaluating alignment only by its effect on classification would overlook its contribution to the review of individual assessments.

A correct prediction does not show that its evidence came from the surveyed house. In more than one-third of correctly aligned images, more than half of the top 10% of activated pixels lay outside the target. In a frame with several houses, a heatmap alone does not tell a reviewer whether a highlighted region belongs to the surveyed house or to a neighboring house. Once the target is outlined in the same image, this becomes a direct visual check. In Figure 9, the outline shows that the strongest activation behind a correct vacant prediction lay on a neighboring building, so the prediction says little about the condition of the surveyed house itself, as Zou and Wang (2021) also suggested. Scores and heatmaps are best read against the designated target house.

Parcel-level aggregation of predictions from several images does not establish that the same building is represented across views, because one image may show the surveyed house while another shows a neighboring house. One possible explanation is that spatial clustering of vacancy keeps such mixed views from lowering parcel-level scores (Mallach, 2018; Morckel, 2014), because a neighboring house often shares the target's occupancy status. If the neighboring building in Figure 9 shares the target's status, the same mechanism could explain why the prediction was correct. The distinction between the target and its neighbors matters, however, when appearance and occupancy do not readily agree, as with an occupied house showing signs of neglect or a well-maintained vacant house, and such cases call for closer examination of the surveyed house and its grounds.

### 5.4. Limitations and future work

The pipeline requires parcel boundaries and panorama capture metadata, which are not available everywhere, and its target selection depends on the quality of the building instance masks. The labels it links to images come from administrative records, and in the vacant-house application the capture date did not always coincide with the survey date, so a label and an image could describe a building in different states.

The classification experiment used a purposive sample that was not nationally representative. Its negatives were not verified in the field, so some unregistered vacant houses may have been among them. Training used one negative per positive, and a candidate pool with a different class share would require recalibrating the threshold and expected performance. The mask-only control reached a parcel-level AUROC close to that of Raw SVI (Table 3). A target mask given to a model can therefore carry class information through its shape and size alone, and studies that supply such masks should report a control of this kind.

The pipeline was developed on panorama images from two Korean platforms. Without target-house labels, the exploratory Detroit run could not validate performance. Beyond vacant houses, the same pipeline could be used to examine the visual features of other building types, provided the filtering criteria are matched to the visual information each task requires. Target selection could also be improved, for example by re-ranking candidate masks according to their convergence across viewpoints or by adding building footprints where they exist, which would matter most in scenes with many candidate buildings. Whether parcels left without a usable image are clustered in particular regions or residential environments remains to be tested. In the longer term, images refined through this pipeline are expected to provide base data for remote building assessment and for secondary studies of individual buildings.

## 6. Conclusion

The SVI available for a parcel does not always show the surveyed building clearly enough to assess it. This study therefore developed a reproducible SVI processing pipeline for remote building assessment and applied it to vacant house assessment in South Korea. The pipeline comprises three procedures. Task-suitability filtering selects images containing readable evidence, target-house alignment connects the building visible in each image to the surveyed parcel, and a classification experiment examines how these two procedures affect classification performance and the visual evidence supporting it.

Task suitability varied with image composition, and group-specific filtering prevented the excessive exclusion of certain operating groups that a single criterion produced. Alignment based on the parcel boundary and capture metadata selected the target house more consistently than frame-based criteria, and selections from different viewpoints converged on the same house. In the classification experiment, marking another house gave nearly the same scores as marking the target, so the scores alone did not show whether the marked house was the surveyed one. Grad-CAM activations appeared on both the target house and the surrounding scene, and target-house alignment provided a reference for reviewing the activated regions against a specific building.

This study advances SVI-based building assessment by showing that image suitability, target identification, and model evidence are separate problems. Parcel-level SVI analysis therefore requires verification not only of final classification performance, but also of whether that performance rests on readable images and the correct target house. For each parcel, the pipeline records which images passed task-suitability filtering and which house in each image was selected as the target, so a reviewer can check both directly. It can thus support human virtual audits (Badland et al., 2010) of individual buildings by designating the house to be examined.

## Declaration of generative AI and AI-assisted technologies in the manuscript preparation process

During the preparation of this work the authors used Claude (Anthropic) in order to proofread the text and improve its readability. After using this tool/service, the authors reviewed and edited the content as needed and take full responsibility for the content of the published article.

# Appendix A. Application to Detroit vacant house data using GSV

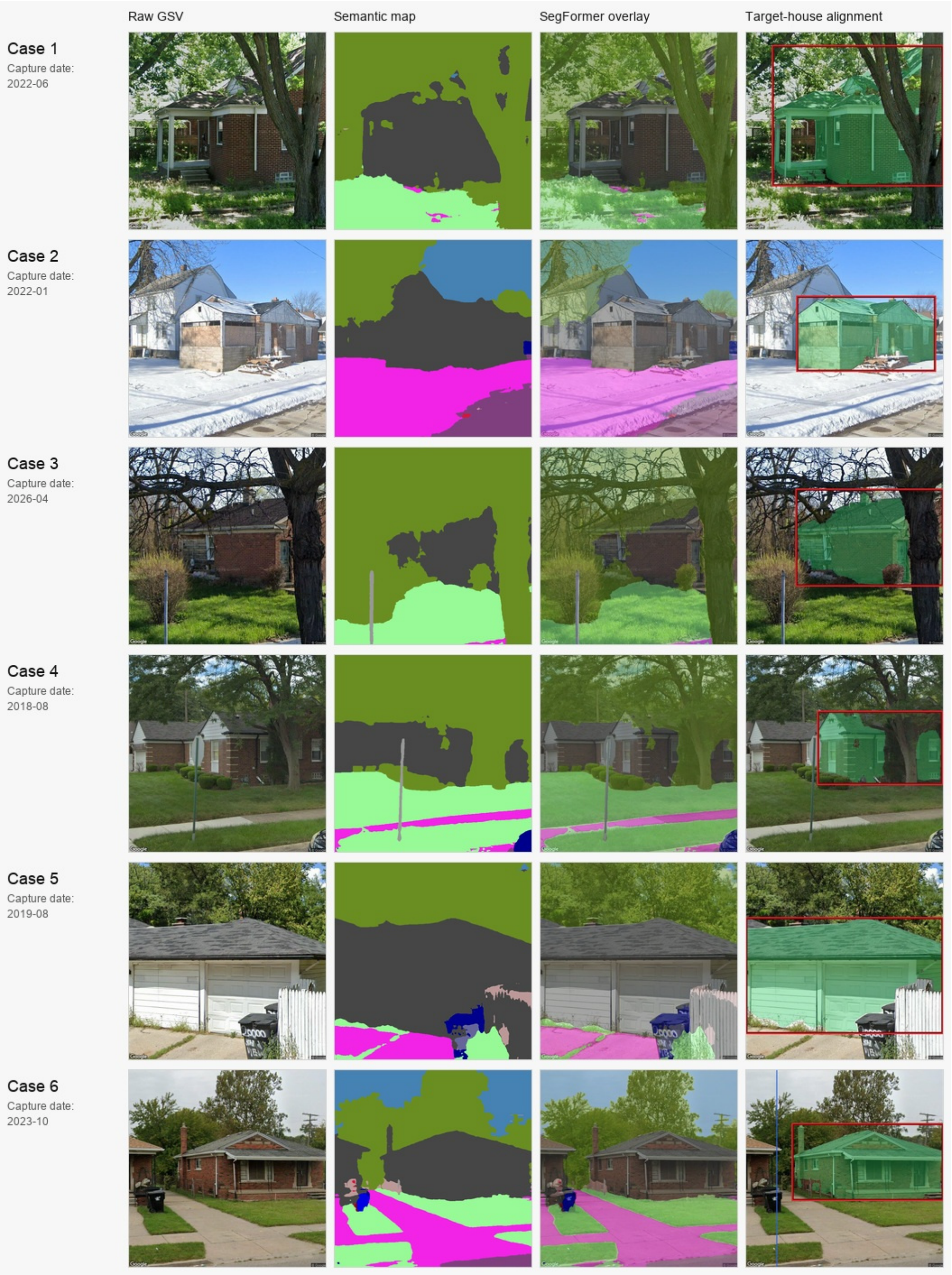

**Figure A.1**. Procedure transferability to Google Street View (Detroit, USA).

To determine whether the proposed pipeline depends on Korea's commercial SVI platforms and cadastral system, we applied it to Detroit, USA. We used as vacancy labels the 30,806 one- to four-unit residential properties classified as probably vacant among the 343,849 records in the Detroit Residential Parcel Survey (DRPS), which was based on a field survey conducted in September 2009, and sampled 5,000 of these parcels. The Google Street View (GSV) Static API replaced Naver Map and Kakao Map as the image source, while City of Detroit parcel polygons replaced the Continuous Cadastral Map. All other parts of the pipeline remained unchanged. Of the 5,000 sampled parcels, 4,984 were matched to parcel polygons, and after locations without a nearby panorama were excluded, the resulting dataset contained 19,612 GSV images.

We processed the images using the same SegFormer segmentation and environmental feature extraction described in the main text. Across the full image set, the mean ratios were 0.399 for vegetation, 0.192 for buildings, and 0.063 for sky, a distribution that differed from the Korean sample and reflected Detroit's low-density residential areas with abundant vegetation. Applying the same filters to each environmental feature group retained 9,273 images (47.3%). Insufficient building and central-building ratios were the main reasons for exclusion, consistent with the area's low-density setting. For the images that passed, SAM 3-based instance mask extraction and target-mask scoring produced a top-1 mask in 8,960 images, after excluding 218 images with no detected mask and 95 in which the parcel was behind the capture position. At the parcel level, multi-view top-1 masks were available for 3,482 parcels. The complete sequence from data acquisition to target-house alignment could therefore be run without changing the pipeline described in the main text.

Because the Detroit sample contains no human-labeled target-house masks, we evaluated the alignment results using an automated check that projected the bottom pixels of each top-1 mask onto the ground and determined whether they fell within the target parcel polygon. By this criterion, 34.0% of valid top-1 images fell within the target parcel. Of the 3,482 parcels, 1,501 (43.1%) had at least 30% of their valid viewpoints fall within the target parcel, and 504 parcels had at least 70%. Figure A.1 shows the six parcels with the highest target-parcel match rates in this set, each of which had four or more valid viewpoints and a target-parcel match for every viewpoint.

Reproducing the pipeline with GSV revealed four problems. (1) Demolition, loss, and reconstruction between the 2009 survey and the GSV capture dates (2018–2026) are not reflected. (2) The current parcel boundaries used for validation may differ from those in place at the time of the survey. (3) Without labeled data, it is impossible to distinguish cases in which a mask located in an adjacent parcel nevertheless represents the correct building. (4) Unlike Korean vacant house inventories, which include a survey date where available, the DRPS gives no survey date for individual houses, making it difficult to assume that the survey and GSV record the same state. This check therefore does not reproduce alignment performance. Instead, it demonstrates that the proposed procedure can be carried out with the same structure on a different platform and in a different urban context. Reaching the conclusions of the main study, however, would require (1) vacant-parcel data recording the survey date for each house, (2) GSV panoramas captured close to those dates, and (3) data confirming whether the house shown in a panorama is the one that was actually surveyed.

# Appendix B. Basis for setting the BCR threshold

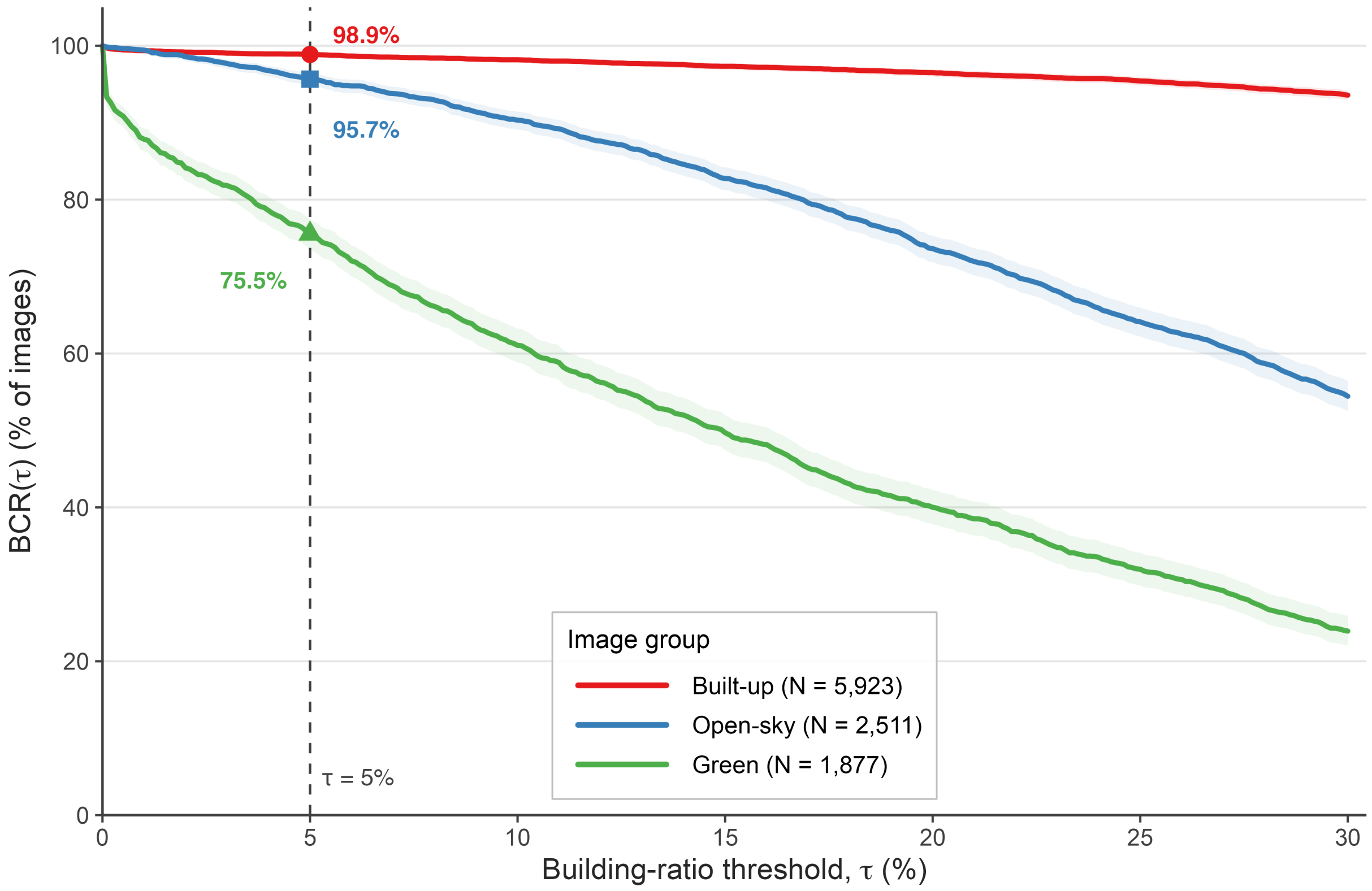

**Figure B.1**. Building Capture Rate (BCR) across building-ratio thresholds for the three image groups.

We selected 5% as the principal threshold for determining whether a building had been detected, balancing the risk that a 1–3% criterion would count minor segmentation errors as detections against the risk that a 10% criterion would exclude images in which a building was present. In the observed data (Figure B.1), the BCR at $\tau = 5\%$ was 98.9% for Built-up, 95.7% for Open-sky, and 75.5% for Green. The BCR remained high where building detection was stable while still revealing the decline observed in the Green group. By contrast, raising the threshold to 10% reduced the BCR for Green to 61.1%, making the effect of the threshold itself disproportionately large relative to the differences among groups. The 5% threshold should therefore be understood as an operational compromise rather than an absolute cutoff. It requires a substantive level of detection while allowing differences among groups to be compared. Robustness was assessed by also reporting results at 1%, 3%, and 10%.

# Appendix C. Detailed filtering criteria and their content

The criteria used to identify unsuitable images draw on two previous studies. The first identified occlusion, capture settings, field of view (FOV), and camera heading as key dimensions of SVI quality (Hou & Biljecki, 2022). The second showed that occlusion, target size, and viewpoint affect object detection performance (Hoiem et al., 2012). For vacant house assessment, we translated these factors into six filters, namely distance, vegetation occlusion, close-up capture, vehicle occlusion, abnormal angle, and central building. The distance, vegetation occlusion, vehicle occlusion, and abnormal angle filters assess capture conditions. The close-up capture and central building filters use the building ratio to determine whether the target building is visible at a scale suitable for assessment. Applying identical filters and thresholds to every image can remove valid images excessively in some environments while allowing unsuitable images to pass in others. We therefore varied both the filters and their thresholds by group to address the main sources of quality degradation in each operating group.

**Table C.1**. Details of the filters.

| Filter | Description | Threshold in prior studies | Pass criterion in this study |
|---|---|---|---|
| Distance filter | Excludes low-quality images in which the panorama is more than a specified distance from the parcel. | Building present within 30 m / 30 m ray | Panorama-to-parcel distance ≤ 30 m |
| Vegetation occlusion filter | Excludes images in which trees or grass obscure too much of the frame. | Vegetation ratio ≤ 0.75 | Vegetation ratio ≤ 0.50 |
| Close-up capture filter | Excludes images in which the building fills so much of the frame that its overall form is difficult to see. | Façades within 5 m and FOV < 50° excluded | Building ratio ≤ 0.85–0.90 |
| Vehicle occlusion filter | Excludes images in which vehicles obscure much of the building. | More vehicle pixels associated with greater error | Vehicle-to-building overlap ratio < 0.02 |
| Abnormal angle filter | Excludes images captured at an abnormal angle that prevents the principal elements from being represented evenly. | Only angles of view (AOV) of 10°–120° allowed | Sky ≥ 0.01, road ≥ 0.01, or building ratio ≥ 0.50 |
| Central building filter | Excludes images in which too little of a building appears in the center of the frame. | Images with a building coverage ratio < 20% excluded | Building ratio in the center (10% of the frame) ≥ 0.03–0.15 |

The filter thresholds were not transferred directly from the literature as a single set of fixed values. Instead, they were informed by operational criteria used in SVI-based building and façade studies, including observation distance, angle of view, building coverage ratio, vegetation ratio, and occluding elements. Previous studies have screened images using an observation radius of approximately 30 m, angle-of-view restrictions of about 10°–120°, a minimum building coverage ratio, and an upper limit on the vegetation ratio (Çelik & Sümer, 2020; Fan et al., 2025; Lee et al., 2015; Liang et al., 2025). Because this study retained multiple observations of each parcel from several platforms, viewpoints, and headings rather than selecting only the single nearest façade image, we applied the earlier criteria conservatively.

The distance filter reflects the loss of detail that occurs as the capture point moves farther from the target parcel, causing exterior cues to appear at a smaller scale and making the house harder to distinguish from adjacent buildings. In extracting façades from GSV panoramas and cadastral maps, Lee et al. (2015) retained only buildings intersected by a raycast within a 30 m buffer, while OpenFACADES also treated the presence of a building within 30 m as an observability criterion when screening candidate images (Lee et al., 2015; Liang et al., 2025). Based on these criteria, we excluded images captured more than 30 m from the parcel.

The vegetation occlusion filter removes cases in which trees and shrubs conceal exterior cues needed to assess vacancy, including exterior walls, entrances, windows, and yards. OpenFACADES retained images with vegetation ratios no greater than 0.75, while Wang (2025) classified images in which trees, boundary walls, or similar objects obscured most of the façade as unusable (Liang et al., 2025; Wang, 2025).

The close-up capture filter removes images in which a building occupies so much of the frame that its overall exterior form, entrance, and surrounding context are cut off. To limit excessive perspective distortion, Lee et al. (2015) excluded façades within 5 m of the camera, those with a minimum FOV below 50°, and those deviating by more than 10° from the camera heading. This study instead used the segmented building ratio directly and classified images with a building ratio above 0.85–0.90 as close-up captures.

The vehicle occlusion filter removes cases in which vehicles conceal cues such as the lower part of the target house, its entrance, windows, and boundary wall. Previous SVI studies have treated external objects, including vehicles, as factors that reduce image usability, and a higher car-pixel count has been associated with greater error in SVI-based estimation (Chen et al., 2026). We excluded images with a vehicle-to-building overlap ratio of 0.02 or greater.

The abnormal angle filter removes images that cannot reasonably be treated as ordinary street-level views. Previous studies have screened out extreme viewing angles using camera parameters such as FOV, AOV, and heading deviation (Lee et al., 2015; Liang et al., 2025). Because platforms do not provide camera parameters in a consistent format, we used segmentation results to classify images in which both sky and road were nearly absent as abnormal views. Images with a building ratio of 0.50 or greater were exempted because they could still provide a sufficient view of the target exterior.

Finally, the central building filter was defined operationally to remove observations in which the target house appears ambiguously at the edge of the frame or an adjacent building occupies the center. Earlier work excluded images with a building coverage ratio below 20% and showed that positional error and occlusion in dense residential areas can cause target buildings to be confused with adjacent buildings (Çelik & Sümer, 2020; Roussel et al., 2024).

**Table C.2**. Criteria for images excluded at the SAM 3 stage.

| # | Reason | Condition | Meaning |
|---|---|---|---|
| 1 | no_house | 0 house masks detected | No house found in the image |
| 2 | too_many | > 8 house masks | Overly dense scene with too many houses |
| 3 | fence_dominant | Boundary wall covers > 50% of a house mask | Boundary wall occludes the house |
| 4 | distant_small | Avg. house mask ratio < 3% or total house pixel ratio < 5% | House is too small or too far away |

# Appendix D. Details of the multi-view consistency analysis

For each image $i$, we calculate the absolute azimuth $\theta_i$ of the top-1 mask center from the panorama position $C_i$, camera heading $\psi_i$, horizontal field of view $\alpha$, image width $W$, and center pixel $x_i$ of the top-1 mask.

$$\theta_i = \psi_i + \left(\frac{x_i}{W} - 0.5\right) \cdot \alpha \qquad \text{(D.1)}$$

We define the ray originating at $C_i$ and extending in direction $\theta_i$ as the sight line $L_i$. If the $N$ sight lines $L_i$ associated with the same parcel point toward a single location, that location can be represented by the point that minimizes the sum of squared perpendicular distances to all $L_i$. We refer to this point as the estimated convergence point $P^*$.

$$P^* = \operatorname{argmin}_P \sum_i d(P, L_i)^2 \qquad \text{(D.2)}$$

When the top-1 masks for a parcel identify the same object, the $L_i$ converge tightly near $P^*$. When different buildings are selected from different viewpoints, the $L_i$ diverge and their distances from $P^*$ increase. We use this distribution of distances to diagnose whether alignment remains consistent across the viewpoints associated with a parcel. A small residual, however, indicates only agreement across viewpoints and does not guarantee that the estimated location corresponds to the input parcel, so the distance from the convergence point to the parcel coordinates should also be considered. The parcel centroid, defined as the center of the polygon, and the convergence point, which represents the building location estimated from the images, refer to different objects. Because the parcel centroid cannot be assumed to coincide with the building centroid, we used 10 m as the threshold for correspondence between the two points (Zandbergen, 2008). In addition, because two viewpoints produce residuals close to zero unless their sight lines are parallel, we evaluated consistency only for parcels with three or more viewpoints.

# Appendix E. Comparison with satellite data

We evaluated the image groups against external data that had not been used in the clustering. Specifically, we compared the green cover percentage derived from satellite NDVI around each parcel with the vegetation ratio estimated by SegFormer to determine whether the vegetation signal in SVI reflected the observed ground environment. To test whether the three image groups represented differences in actual ground-level greenery, we compared the image-based vegetation ratio with the Google Earth Engine (GEE) green cover percentage at the same capture coordinates (Figure E.1). The correlation was $r = 0.530$ ($p < 0.001$). This result indicates that the three image groups reflect differences in the physical green environment rather than an arbitrary grouping.

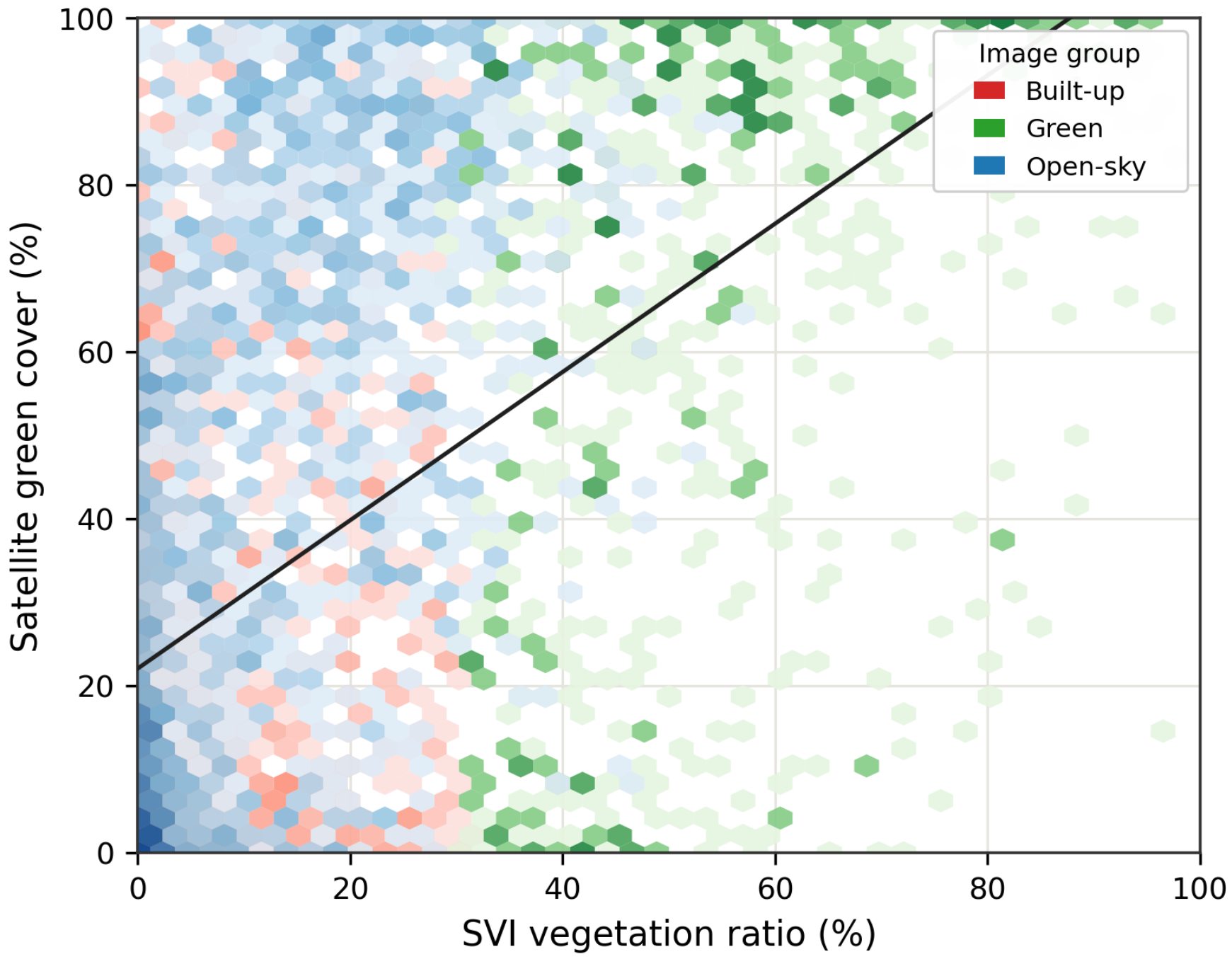

**Figure E.1**. Comparison with NDVI green cover percentage.